# Qualitative Models for Decision Under Uncertainty without the Commensurability Assumption

**Hélène Fargier**
IRIT - Paul Sabatier University
118 route de Narbonne
31062 Toulouse Cedex 4, France

**Patrice Perny**
LIP6 - Paris 6 University
4 Place Jussieu
75252 Paris Cedex 05, France

## Abstract

This paper investigates a purely qualitative version of Savage's theory for decision making under uncertainty. Until now, most representation theorems for preference over acts rely on a numerical representation of utility and uncertainty where utility and uncertainty are commensurate. Disrupting the tradition, we relax this assumption and introduce a purely ordinal axiom requiring that the Decision Maker (DM) preference between two acts only depends on the relative position of their consequences for each state. Within this qualitative framework, we determine the only possible form of the decision rule and investigate some instances compatible with the transitivity of the strict preference.

## 1 INTRODUCTION

In the field of decision making under uncertainty, several studies have been carried out to justify the use of this or that decision criterion for the comparison of alternatives. Each decision criterion is justified by a set of axioms capturing part of the decision behaviour of the DM. Among the most classical studies in this direction are the seminal works of Von Neumann and Morgenstern (1947) and Savage (1954). Such approaches rely on the use of a quantitative criterion for the comparison of alternatives whose justification requires several strong assumptions. For instance, the axiom system usually implies that the subjective value attached to each consequence can be quantified, as well as the likelihood of the possible events. However, in practical applications, the elicitation of the information required by a quantitative model cannot always be assessed. This is why some alternative models have been proposed in AI, relying on a more ordinal representation of preferences and uncertainty (*e.g* the *qualitative utility* proposed by Dubois and Prade (1995)).

Even if qualitative utility mainly relies on the use of ordinal information (preference order on consequences, relative likelihood of events) it shares a common feature with the expected utility criterion: *the commensurability of the preference scale and the uncertainty scale used in the model*. Recent works in AI propose to escape this assumption in two different ways. The first one is to compare two acts on the basis of their consequences in the *most plausible states of the world* (Boutilier (1994); Tan and Pearl (1994); Brafman and Tennenholtz (1996), (1997)). The other way is to rely on a decision rule that compares the plausibility of the sets of states in which one act has a better consequence than the other (see Dubois et al. (1997)). This rule has been originally proposed in a possibilitic framework, but could be easily extended to other uncertainty relations, thus defining other preference models. The aim of the present paper is to provide an unified axiomatic framework for these kind of rules and to study their ability to describe the DM preferences.

In the next section, we present various quantitative and qualitative models for decision making under uncertainty and discuss the necessity of making preference and uncertainty commensurate. In Section 3, an axiomatic framework for purely qualitative preference models is introduced and the structure of qualitative decision rules is characterized. In Section 4 we characterize the qualitative decision rules compatible with the transitivity of the strict preference. Finally, in Section 5, we discuss some variations of our framework and propose a new family of qualitative decision rules compatible with the transitivity requirement. Notice that in order to save space, all the proofs are omitted.

## 2 EXAMPLES AND MOTIVATIONS

Decision Making under uncertainty implies a choice among a set of potential acts (decisions) whose consequences are not perfectly known. From a formal point of view, such a decision problem is characterized by a set $S$ of states representing the possible situations (or decision contexts), a set $X$ of possible consequences, and a set of acts viewed as elements of $X^S$ (*i.e.* an act is a mapping $f : S \to X$ where $f(s)$ represents the consequence of act $f$ for any state $s \in S$). In this paper, $S$ and $X$ are supposed to be finite[1]. Thus, if $X = \{x_1, \ldots, x_m\}$ and $S = \{s_1, \ldots, s_n\}$, an act $f \in X^S$ is completely characterized by the vector of consequences $(f(s_1), \ldots, f(s_n))$

Denoting $\succsim$ the DM's preference relation over acts, we define the *indifference relation* $\sim$, and a *strict preference relation* $\succ$ by:

$$f \sim g \iff f \succsim g \text{ and } g \succsim f$$
$$f \succ g \iff f \succsim g \text{ and not}(g \succsim f)$$

The role of decision analysis is to construct a model of DM's preferences $\succsim$, including his subjectivity in the evaluation of possible events and consequences of alternatives, so as to be able to use this model to support the decision making process. Several preference models have been proposed in this direction, that are all based on a decision rule of the following form:

$$f \succsim g \iff \chi(f(s_1), \ldots, f(s_n), g(s_1), \ldots, g(s_n)) \quad (1)$$

where $\chi$ is a comparison function defined from $X^{2n}$ to $\{0, 1\}$ (0 for 'false' and 1 for 'true'). Now we distinguish two types of models corresponding to two different instances of equation (1).

### 2.1 PREFERENCE MODELS BASED ON A VALUE FUNCTION

These models are characterized by the definition of a value function $v : X^S \to Y$ where $Y$ is an ordered set and $v(f)$ measures the subjective attractiveness of $f$ for the DM. In such models, $v(f) = \psi(f(s_1), \ldots, f(s_n))$ where $\psi$ is an aggregation function defined from $X^n$ to $Y$. These models are characterized by the following instance of equation (1)

$$f \succsim g \iff \varphi(\psi(f(s_1), \ldots, f(s_n)), \psi(g(s_1), \ldots, g(s_n)))$$

where $\varphi$ is a comparison function defined from $Y^2$ to $\{0, 1\}$. Let us recall some well known examples:

#### 2.1.1 The Expected Utility

This model is based on a utility function $u : X \mapsto [0, 1]$ measuring the subjective attractiveness of any consequence $x \in X$ and a probability distribution $p$ on $S$ defined by the given of the vector $(p(s_1), \ldots, p(s_n))$. In this probabilistic model

---

[1]This differs from Savage's original framework for decision making under uncertainty

$\psi(z_1, \ldots, z_n) = \sum_{i=1}^n p(s_i)u(z_i)$. This defines a criterion $v(f)$ for any $f \in X^S$ by:

$$v(f) = \sum_{s \in S} p(s)u(f(s)) \quad (2)$$

#### 2.1.2 The Qualitative Utilities

The pessimistic qualitative utility (see *e.g.* Dubois and Prade (1995); Dubois et al. (1998)) is based on a utility function $u : X \mapsto [0, 1]$ measuring the attractiveness of a consequence $x \in X$ and a possibility distribution $\pi$ on $S$ defined by the vector $(\pi(s_1), \ldots, \pi(s_n))$. In this model $\psi(z_1, \ldots, z_n) = \min_{i=1,\ldots,n} \max\{1 - \pi(s_i), u(z_i)\}$. This defines the pessimistic criterion $v^-(f)$ for any $f \in X^S$ by:

$$v^-(f) = \min_{s \in S} \max\{1 - \pi(s), u(f(s))\} \quad (3)$$

This criterion is actually a *weighted* extension of the Wald criterion and, to some extent, it is "more qualitative" than the standard expected utility.

The optimistic qualitative utility model (see *e.g.*; Dubois and Prade (1995); Dubois, et al (1998)) is a variation of the previous model, and as such, it requires the same utility function and the same possibility distribution. This model is characterized by $\psi(z_1, \ldots, z_n) = \max_{i=1,\ldots,n} \min\{\pi(s_i), u(z_i)\}$. This defines the optimistic qualitative criterion $v^+(f)$:

$$v^+(f) = \max_{s \in S} \min\{\pi(s), u(f(s))\} \quad (4)$$

### 2.2 THE COMMENSURABILITY HYPOTHESIS

Let us underline some implicit assumptions in decision models based on value functions, namely the commensurability between preference and uncertainty, and, for the expected utility model, the cardinality of utilities. Suppose that the DM's preference over $X$ are known and can be represented by the weak-order $\succsim_P$ ($\succsim_P$ is precisely the order encoded by the utility function). Similarly, suppose that the subjective likelihood of each event for the DM is known and is described by a binary uncertainty relation $\succsim_U$ on $2^S$ (e.g. represented by the probability/possibility/necessity measure).

**Example 1** *Let us consider a decision problem with two states $\{s_1, s_2\}$ which are seen as equally plausible by the decision maker (i.e. $s_1 \succsim_U s_2$ and $s_2 \succsim_U s_1$) and a set of consequences $X = \{x_1, x_2, x_3, x_4\}$ such that $x_1 \succ_P x_2 \succ_P x_3 \succ_P x_4$. Let us consider two different utility functions $u_1$ and $u_2$ that could be used to encode the preference order $\succ_P$:*

|       | $x_1$ | $x_2$ | $x_3$ | $x_4$ |
|-------|-------|-------|-------|-------|
| $u_1$ | 0.9   | 0.6   | 0.4   | 0.3   |
| $u_2$ | 0.7   | 0.6   | 0.4   | 0.1   |

*In this paper, we denote $f_1A_1f_2A_2,\ldots,A_{n-1}f_nA_n$ the act whose result is $f_i(s)$ if $s \in A_i, \forall i = 1,\ldots,n$, for any partition $A_1, \ldots, A_{n-1}, A_n$ of the set of states. Consider the acts $f = x_1\{s_1\}x_4\{s_2\}$ and $g = x_2\{s_1\}x_3\{s_2\}$. Suppose now that we want to use the expected utility model to compare $f$ and $g$. The probability distribution on $S$ must be $p(s_1) = p(s_2) = 0.5$. Using function $u_1$ in equation (2) we get the expected values $v_1(f) = 0.6$, $v_1(g) = 0.5$. Performing the same operations with $u_2$ yields to $v_2(f) = 0.4$, $v_2(g) = 0.5$. Therefore we get $f \succ g$ with $u_1$ and $g \succ f$ with $u_2$.*

Hence, when using expected utility, the preference over acts depends on the particular utility function chosen to encode the preference order $\succ_P$. This shows that this model exploits an extra information not contained in relations $\succsim_P$ and $\succsim_U$. This is due to the cardinal interpretation of utility grades and to their interlacement with the probabilities.

In this example, whenever we use a possibility distribution $\pi$ to describe our knowledge about the plausibility of events, we have to set: $\pi(s_1) = \pi(s_2) = 1$. Then, the pessimistic qualitative utility gives: $v_1^-(f) = 0.3$ and the $v_1^-(g) = 0.4$ from $u_1$ and therefore $g \succ f$, and we get exactly the same preference with function $u_2$ since $v_2^-(f) = 0.1$ and the $v_2^-(g) = 0.4$. Thus, on this example where the initial information (represented by relations $\succsim_P$ and $\succsim_U$) is purely ordinal, using qualitative utility seems more natural than expected utility. From such an example, we could infer that this ordinal model is purely qualitative. However, things are less elementary, as shown in the following example.

**Example 2** *Suppose now that $s_1 \succ_U s_2$ in example 1 and consider the following utility functions:*

|     | $x_1$ | $x_2$ | $x_3$ | $x_4$ |
|-----|-------|-------|-------|-------|
| $u_1$ | 0.4 | 0.3 | 0.2 | 0.1 |
| $u_2$ | 0.9 | 0.8 | 0.7 | 0.6 |

*We want now to compare the acts $f$ and $g$ on the basis of the pessimistic qualitative utility model. Let us choose an arbitrary possibility distribution on $S$ to represent the likelihood weak-order on $\succsim_U$, e.g $\pi(s_1) = 1$ and $\pi(s_2) = 0.5$. Hence, using function $u_1$ in equation (3) we get the expected values $v_1^-(f) = 0.4$, $v_1^-(g) = 0.3$. Performing the same operations with function $u_2$ yields to $v_2^-(f) = 0.6$, $v_2^-(g) = 0.7$. Therefore we get $f \succ g$ with $v_1^-$ and $g \succ f$ with $v_2^-$.*

This example shows that models based on qualitative utility also exploit an extra information not contained in relations $\succsim_P$ and $\succsim_U$, namely their interlacement. Indeed, in such models, utility and possibility are commensurate and represented on a common ordinal scale $PU$. On this scale, the use of function $u_1$ yields to the following order: $s_1 \succ_{PU} s_2 \succ_{PU} x_1 \succ_{PU} x_2 \succ_{PU} x_3 \succ_{PU} x_4$ whereas function $u_2$ yields to: $s_1 \succ_{PU} x_1 \succ_{PU} x_2 \succ_{PU} x_3 \succ_{PU} x_4 \succ_{PU} s_2$. This explains the difference of result. Constructing relation $\succ_{PU}$ from $\succ_U$ and $\succ_P$ requires additional information which must be elicited by questioning the DM. In the following Section, we introduce simple models avoiding the definition of relation $\succsim_{PU}$.

### 2.3 PREFERENCE MODELS BASED ON PAIRWISE COMPARISONS

We consider here models characterized by the following general instance of equation (1).

$$f \succsim g \iff \varphi'[\psi[\varphi(f(s_1),g(s_1)),\ldots,\varphi(f(s_n),g(s_n))], \\ \psi[\varphi(g(s_1),f(s_1)),\ldots,\varphi(g(s_n),f(s_n))]]$$

where $\varphi$ and $\varphi'$ are comparison functions and $\psi$ is an aggregation function. When the preferences over $X$ are represented by the weak-order $\succsim_P$ we can choose:

$$\varphi(x_1,x_2) = \begin{cases} 1 & \text{if} \quad x_1 \succsim_P x_2 \\ 0 & \text{otherwise.} \end{cases}$$

Now, if we assume that the subjective likelihood of each event for the DM are described by a binary uncertainty relation $\succsim_U$ on $2^S$, then for any pair of events $A, B$ seen as elements of $2^S$ and characterized by vectors $(a_1,\ldots,a_n)$ and $(b_1,\ldots,b_n)$ respectively, we can define function $\psi$ so as to satisfy the condition:

$$\psi(a_1,\ldots,a_n) \geq \psi(b_1,\ldots,b_n) \iff A \succsim_U B$$

This yields to the following general decision rule:

$$f \succsim g \iff [f \succsim_P g] \succsim_U [g \succsim_P f] \qquad (5)$$

where $[f \succsim_P g] = \{s \in S, f(s) \succsim_P g(s)\}$. This general *Lifting Rule*, proposed by Dubois and al. (1997), can be seen as the counterpart of majority rules in social choice theory (see *e.g* Bordes (1983)).

The *Probabilistic Lifting rule*, can be represented by a function $\psi$ defined by:

$$\psi(a_1,\ldots,a_n) = \sum_{\{j:a_j=1\}} p(s_j)$$

where $p$ is a probability distribution on $S$.

The *Necessity Lifting rule*, corresponds to :

$$\psi(a_1,\ldots,a_n) = \min_{\{j:a_j=0\}} (1 - \pi(s_j))$$

where $\pi$ is a possibility distribution on $S$. Symmetrically, we can propose a new *Possibilistic Lifting Rule*, represented by a function $\psi$ defined by:

$$\psi(a_1,\ldots,a_n) = \max_{\{j:a_j=1\}} \pi(s_j)$$

Possibilistic and necessity lifting rules yield to a relation $\succsim$ whose asymmetric part is transitive but that this is not the case for the probabilistic lifting. These three rules yield to a complete relation $\succsim$, however we can consider more general lifting rules yielding to incomplete preference structures. For example, one can consider a family of possibility distributions, and state that $f$ is preferred to $g$ iff, for each possibility distribution in the family, the possibility of $[f \succsim_P g]$ is greater than the possibility of $[g \succsim_P f]$.

Finally, notice that the preference inversions observed in example 1 and 2 cannot occur when using lifting rules because, by definition, the preference over acts only depends on relations $\succsim_U$ and $\succsim_P$. In this respect, preference models based on a lifting rule can be seen as *purely qualitative* models.

## 3 A QUALITATIVE VERSION OF SAVAGE's FRAMEWORK

This Section aims at providing an axiomatic framework for qualitative preference models and to derive representation theorems for purely qualitative preferences. Following Savage we start from a user-driven preference relation over acts, and define axioms that this preference relation should satisfy to be a qualitative and "rational" model. Under such axioms, it is possible to construct a preference relation $\succsim_P$ on the consequences, a likelihood relation $\succsim_L$ on events, and a decision rule based on these two relations that perfectly describes the DMs preferences. Let us first recall a part of the axioms proposed by Savage (1954) to characterize the expected utility model (2).

### 3.1 SAVAGE's AXIOMS

**Axiom S1.** $(X^S, \succsim)$ is a weakly ordered space ( $\succsim$ is complete, reflexive, transitive)

Such an axiom in Savage's work is justified by the goal assigned to the theory. If acts are ranked according to expected utility then the preference over acts will be transitive, reflexive, and complete. In this paper, we do not want to require such a property a priori: the DM Preferences could be rational without being complete nor transitive (see Section 4).

Now, let $A \subseteq S$ be an event, $f$ and $g$ two acts, and denote by $fAg$ the act such that:

$$fAg(s) = \begin{cases} f(s) & if \quad s \in A \\ g(s) & if \quad s \notin A \end{cases}$$

$fAg$ is short for $fAg\bar{A}$ where $\bar{A}$ is the complement of $A$. Savage has proposed the sure-thing principle axiom, that requires that the preference between two acts $fAh$ and $gAh$ does not depend on the choice of $h$:

**Axiom S2 (Sure-thing principle).** $\forall A \subseteq S$, $\forall f, g, h, h' \in X^S, (fAh \succsim gAh \iff fAh' \succsim gAh')$

**Definition 1 (Conditional Preference)** $f$ is said to be weakly preferred to $g$, conditioned on $A$ iff if $\forall h \in X^S, fAh \succsim gAh$. This is denoted by $(f \succsim g)_A$.

**Definition 2 (Null events)** An event $A$ is said to be null if and only if: $\forall f, g, h \in X^S : fAh \sim gAh$.

Any event $A \subseteq S$ unable to make a discrimination for at least one pair of acts is considered as null (e.g irrelevant for the DM). Note that this definition of null events is also consistent when $\succsim$ is not complete.

Among acts in $X^S$ there are *constant acts* such that: $\exists x \in X : \forall s \in S, f(s) = x$. Such an act will be denoted $f_x$. It seems reasonable to identify the set of constant acts $\{f_x, x \in X\}$ and $X$. The DM preference relation $\succsim_P$ on $X$ can be induced from $(X^S, \succsim)$ as follows:

$$\forall x, y \in X, \quad (x \succsim_P y \iff f_x \succsim f_y) \qquad (6)$$

Another hypothesis is that the preference order on consequences is unique and does not depend on the events considered. This is the third Savage postulate:

**Axiom S3.** $\forall A \subseteq S, A$ not null, $(f_x \succsim f_y)_A \iff x \succsim_P y$.

The preference on acts also induces a likelihood relation among events: it is sufficient to consider the set of binary acts, of the form $f_xAf_y$, which thanks to (S3) can be denoted $xAy$, where $x \succ_P y$. Clearly for fixed $x \succ_P y$, the set of acts $\{x, y\}^S$ is isomorphic to the set of events $2^S$. Since the restrictions of $(X^S, \succsim)$ to $\{x, y\}^S$ may be inconsistent with the restriction to $\{x', y'\}^S$ for other choices of consequences $x', y'$ such that $x' \succ_P y'$, Savage has introduced a new postulate:

**Axiom S4.** $\forall A, B \subseteq S, \forall x, y, x', y' \in X : x \succ_P y$ and $x' \succ_P y', xAy \succsim xBy \iff x'Ay' \succsim x'By'$.
Under S4, the choice of $x, y \in X$ with $x \succ_P y$ is not important when defining the ordering between events in terms of binary acts. Hence, the following likelihood relation on events can be derived from $\succsim$:

$$A \succsim_L B \Leftrightarrow (\exists x, y \in X : x \succ_P y \text{ and } xAy \succsim xBy) \quad (7)$$

Lastly, Savage has assumed that the weakly ordered set $(X, \succsim_P)$ is not trivial:

**Axiom S5.** $X$ contains at least two elements $x, y$ such that $f_x \succ f_y$ (or $x \succ_P y$).
Under S1-S5, the likelihood relation $\succsim_L$ on events is a comparative probability ordering (be $S$ finite or not).

Finally, Savage introduces last postulates that enable him to derive the existence (and uniqueness) of a numerical probability measure on $S$ that can represent the likelihood relation $\succsim_L$. However, these axioms are omitted in the present paper because they are irrelevant for a finite set of states and for qualitative models.

## 3.2 THE QUALITATIVE INDEPENDENCE

We propose now a new axiom aiming at specifying the qualitative nature of a preference model. This axiom is an independence axiom that can be seen as the counterpart of an independence condition used in social choice theory (see Sen (1986)), completed with a neutrality condition making preferences independent of the labels of the acts considered. We call this axiom *Qualitative Independence*:

**Axiom QI.** $\forall h, g, h', g' \in X^S$, $[(h,g) \equiv (h',g')] \Longrightarrow (h \succsim g \iff h' \succsim g')$, where $(h,g) \equiv (h',g')$ stands for $[\forall s \in S, (f_{h(s)} \succsim f_{g(s)} \iff f_{h'(s)} \succsim f_{g'(s)})$ and $(f_{g(s)} \succsim f_{h(s)} \iff f_{g'(s)} \succsim f_{h'(s)})]$.

Note that, under S3, QI can be soundly rewritten as:

**Axiom QI'.** $[\forall s \in S, (h(s) \succsim_P g(s) \iff h'(s) \succsim_P g'(s))$ and $(g(s) \succsim_P h(s) \iff g'(s) \succsim_P h'(s))]$ $\Longrightarrow (h \succsim g \iff h' \succsim g')$

This new version of QI expresses that only the relative positions of the consequences of the two acts are important, but not the consequences themselves, nor the positions of the two acts relatively to other acts (*i.e.* the preference relation is not context dependent). As a consequence, $QI$ holds for preferences derived using the qualitative decision rules introduced in 2.3:

**Proposition 1** *The general lifting rule (5) (whose definition supposes S2, S3 and S4) satisfies QI.*

Actually, when S3 is assumed, QI can be understood as a strong version of S2 and also of S4, as soon as $\succsim$ is supposed to be reflexive. Indeed we have:

**Proposition 2** *(S3 + QI + $\succsim$ reflexive) $\Longrightarrow$ S2*

**Proposition 3** *(S3 + QI + $\succsim$ reflexive) $\Longrightarrow$ S4*

Since S2 and S4 hold, assuming S3 and S5 makes it possible to briefly state the following properties of the projection of $\succsim$ on $S$, *i.e.* $\succsim_L$ as defined in (7):

**Proposition 4** *(S3 + S5 + QI + $\succsim$ reflexive) $\Longrightarrow$*

    *i)*    $S \succ_L \emptyset$,   *ii)*  $\forall A \subseteq S, A \succsim_L \emptyset$
    *iii)*  $\forall A \subseteq S, \emptyset \succsim_L A \iff A$ *is null*
    *iv)*  $B$ *is null and* $A \subseteq B$, *then* $A$ *is null*
    *v)*  $\succsim_L$ *is additive*

**Proposition 5** *If $\succsim$ is complete on constant acts, reflexive, satisfies S3 and QI then there exists a preference relation $\succsim_P$ on $X$ defined by (6) and a likelihood relation $\succsim_L$ on $S$ defined by (7) such that, for any $f, g \in X^S$, $f \succsim g \iff [f \succsim_P g] \succsim_L [g \succsim_P f]$*

This provides a sufficient condition for DM preferences to be representable by a lifting rule. Then, using proposition 1 we get:

**Corollary 1** *If the DM preference $\succsim$ is complete on constant acts, reflexive on $X^S$ and satisfies S3, QI holds if and only if $\succsim$ is a lifting rule.*

This result provides a useful characterization of the lifting rule that can be used to justify this model in a practical situation. If QI is accepted as a norm for qualitative models, then the only available decision rules are lifting rules. This result holds even if $\succsim$ is not supposed to be complete, and thus concerns a wider class of lifting rules than those given in paragraph 2.3 and in Dubois et al. (1997). Under QI, we get nice unanimity properties: whenever $f$ is as least as good $g$ for all possible states, $f$ must be preferred to $g$:

**Proposition 6** *If $\succsim$ satisfies S3, S5 and QI then $\forall f, g \in X^S$, $(\forall s \in S, f(s) \succ_P g(s)) \Longrightarrow f \succ g$*
*$\forall f, g \in X^S, (\forall s \in S, f(s) \succsim_P g(s)) \Longrightarrow f \succsim g$*

*Moreover, if $\succsim$ is complete then we get:*
*$\forall f, g \in X^S, (((\forall s \in S, f(s) \succsim_P g(s))$ and*
*$(\exists s \in S, s$ not null and $f(s) \succ_P g(s))) \Longrightarrow f \succ g)$*

## 4 THE TRANSITIVITY OF STRICT PREFERENCE

The transitivity of preference resulting from a qualitative decision rule is not obvious. This can be simply observed using the following example:

**Example 3** *Suppose $S = \{s_1, s_2, s_3\}$ with probabilities $p(s_1) = p(s_2) = p(s_3) = 1/3$. Consider 3 consequences $x, y, z \in X$ such that $x \succ_P y \succ_P z$. Using the probabilistic lifting to compare the acts $f = x\{s_1\}y\{s_2\}z\{s_3\}, g = y\{s_1\}z\{s_2\}x\{s_3\}$ and $h = z\{s_1\}x\{s_2\}y\{s_3\}$ we get $[f \succsim g] = \{s_1, s_2\}, [g \succsim h] = \{s_1, s_3\}$ and $[h \succsim f] = \{s_2, s_3\})$ and therefore $f \succ g, g \succ h$ and $h \succ f$, i.e. an intransitive cycle. On the contrary if probabilities of states are $p'(s_1) = 0.6, p'(s_2) = 0.3$ and $p'(s_3) = 0.1$ we get $f \succ g, g \succ h$ and $f \succ h$, a complete order.*

This example is actually derived from a Condorcet effect as those observed in Social choice (see *e.g* Sen, 1986)). Let us briefly clarify the relationships with ordinal aggregation, defining for each state $s \in S$ a preference relation $\succsim_s$ by:

$$\forall f, g, \in X^S, \quad f \succsim_s g \iff f(s) \succsim_P g(s) \quad (8)$$

In this context, QI is a condition requiring that preference $f \succsim g$ only depends on partial preferences $f \succsim_s g$ and $g \succsim_s f$, $s \in S$, and not on the consequences $f(s)$ and $g(s)$, $s \in S$ themselves. Therefore, under QI, linking relation $\succsim$ to the profile $(\succsim_{s_1}, \ldots, \succsim_{s_n})$ can be seen as an *ordinal* aggregation problem, comparable to those studied in Social Choice Theory.

From a descriptive point of view, example 3 shows the ability of qualitative decision rules to explain some cyclic preference structures and this is a really original point when compared to more classical models based on value functions. However, from a prescriptive point of view, the possibility of getting cyclic preferences is an important source of difficulty. For this reason, we investigate in this Section the possibility for qualitative models to fulfill some minimal transitivity requirements. To this end, we must introduce new definitions concerning the decisiveness of events.

### 4.1 DECISIVE EVENTS

In example 3, considering the probability distribution $p'$, it is easy to show that $f \succ_P g$ holds as soon as $f(s_1) \succ_P g(s_1)$ and thus, $s_2$ and $s_3$ are negligible with respect to $s_1$. $\{s_1\}$ is said to be a *decisive state*:

**Definition 3** *For any subset $S' \subseteq S$, an event $A \subseteq S'$ is said to be decisive in $S'$ iff: $\forall f, g \in X^S$, $[[(\forall s \in S \setminus S', f(s) \sim_P g(s))$ and $(\forall s \in A, f(s) \succ_P g(s))] \Longrightarrow f \succ g]$. A decisive event in $S$ is called a* decisive event. *A decisive singleton is called a* decisive state.

In example 3, the transitivity is obtained using a very special structure: $\{s_1\}$ is a decisive state, $\{s_2\}$ is a decisive state in $\{s_2, s_3\}$ and this yields to the following hierarchy of decisive events $\{s_1\} \succ_L \{s_2, s_3\} \succ_L \{s_2\} \succ_L \{s_3\}$ which corresponds to a lexicographic preference structure (see Fishburn, 1975). A weakening of decisiveness for states is defined by:

**Definition 4** *For any subset $S' \subseteq S$, the state $s' \in S'$ is said to be a vetoer in $S'$ iff: $\forall f, g \in X^S$, $[[(\forall s \in S \setminus S', f(s) \sim_P g(s))$ and $f(s') \succ_P g(s')] \Longrightarrow not(g \succ f)]]$. A vetoer in $S$ is simply called a* vetoer.

**Definition 5** *For any subset $S' \subseteq S$, an event $A \subseteq S'$ is said to be* predominant *in $S'$ iff it is decisive in $S'$ and contains only vetoer-states in $S'$. A predominant event in $S$ is simply called a* predominant event.

### 4.2 MODELS COMPATIBLE WITH QUASI-TRANSITIVE PREFERENCES

We consider here a very general framework where preferences are not supposed to be transitive nor complete.

We only assume the *quasi-transitivity* of $\succsim$, e.g. the transitivity of its asymmetric part $\succ$. This weak requirement allows a wide class of preference models to be concerned, including those with an intransitive indifference (semi-orders, interval orders, see e.g Pirlot and Vincke (1997)) and those allowing incomparability (see e.g. Roy, (1996)). We require only the total comparability of constants acts (i.e. the completeness of $\succsim_P$) which is a natural property because there is no conflict between states in the comparison of constant acts. Hence we consider the following axiom that can be seen as a useful weakening of S1.

**Axiom A1.** $\succsim$ is reflexive and quasi-transitive, and its restriction to constant acts is complete.

Moreover, as far as transitivity properties are concerned, we need to consider at least three distinct levels on $X$ and strengthen axiom S5 with the following:

**Axiom A5.** $X$ contains at least three consequences $x$, $y$ and $z$ such that $f_x \succ f_y \succ f_z$.

Then we establish the two following technical lemmas:

**Lemma 1** *If $\succsim$ satisfies A1, A5, S3, and QI then the following property holds for any non-empty subset $A$ of $S$: $[\exists f, g \in X^S : ([f \succ_P g] = A, [g \succ_P f] = S \setminus A$ and $f \succ g)] \implies A$ is decisive*

**Lemma 2** *If $\succsim$ satisfies A1, A5, S3, and QI then $S$ contains at most one predominant event.*

Hence we get the following theorem:

**Theorem 1** *If $\succsim$ satisfies A1, A5, S3 and QI, there exists one and only one predominant event $O$ in $S$ and the resulting subjective likelihood relation $\succ_L$ on events has the following properties:*

i) $\forall A, B \subseteq S, (A \cap B = \emptyset$ and $A \succ_L B) \Rightarrow O \cap B = \emptyset$
ii) $\forall A, B \subseteq S, (A \cap B = \emptyset$ and $O \subseteq A) \Rightarrow A \succ_L B$
iii) $\forall A \subseteq S, \quad O \subseteq A \iff A \succ_L S \setminus A$

In other terms, when the DM preferences are quasi-transitive and compatible with QI, they necessarily rely on a knowledge structure admitting a predominant event $O$ which makes negligible any other disjoint event. Indeed $f \succ g$ holds as soon as $f$ is preferred to $g$ for all states belonging to $O$. Moreover, whenever two states $s$ and $s'$ are conflicting within $O$ concerning a pair $f, g$ (i.e. $f(s) \succ_P g(s)$ and $g(s') \succ_P f(s')$), no strict preference can be stated between the two acts. In Social Choice Theory, such a preference structure is said to be *oligarchic* (see e.g. Sen (1986)). The larger is $O$, the less decisive the procedure. In case of total ignorance, i.e. when no state of the world is more likely than another, we get $O = S$: the preference relation $\succsim$ reduces to the *functional dominance* on $X^S$.

In the other extreme case, $O$ contains a single state, the DM preferences can be described by a "dictatorial" decision rule due to the existence of a decisive state $s$. Such a rule can emerge when the DM is almost certain than the "true state" is $s$ and that the other states are negligible: all the other possible futures are ignored. More precisely, we can derive the following result:

**Corollary 2** *If axioms A1, A5, S3, and QI hold, there exists a unique maximal[2] partition of $S$ into events $O_1, \ldots, O_k$, $k > 1$ such that $O_1$ is predominant and $\forall j : 2 \leq j < k$, event $O_j$ is predominant in $S \setminus \bigcup_{i=1}^{j-1} O_i$ thus yielding to a hierarchy of predominant events.*

This result is well illustrated by possibilistic and necessity lifting rules for which A1, A5, S3 and QI hold. The possibilistic lifting corresponds to the case where only one predominant event $O_1$ exists. It is simply defined by the core of the distribution, e.g. the set of states having a maximal possibility: $O_1 = \{s \in S : \pi(s) = 1\}$. The necessity lifting rule corresponds to the case with possibly several levels in the hierarchy, i.e. one for each distinct level-cut of the distribution $\pi$ on $S$.

Now, assuming both completeness and transitivity of $\succsim$ we obtain a hierarchy of predominant *states*:

**Theorem 2** *For any preference relation among acts that satisfies S1, S3, A5 and QI, we have, for all pairs $f, g \in X^S$*
$$f \succ g \iff \begin{cases} \exists s \in S, \text{not null}, \ f(s) \succ_P g(s) \\ \forall s \in S, \ (g(s) \succ_P f(s)) \implies \exists s' \in S : \\ \{s'\} \succ_L \{s\} \text{ and } f(s') \succ_P g(s')) \end{cases}$$

This theorem shows the lexicographic structure of preference and can be seen as a counterpart of Fishburn's theorem obtained in the context of multicriteria analysis (see Fishburn, 1975). It demonstrates a phenomenon which was already illustrated in example 3 and also explains that comparative probabilities (and not only classical probabilities) are incompatible with a qualitative approach of decision. The resulting lexicographic rule corresponds to a very particular structure of uncertainty in which all the non impossible states must be linearly ordered. As a consequence, this forbids the description of total ignorance (where the states are equally plausible). Moreover, there exists a predominant state $s^*$, more plausible that any other state, and even more plausible than the event $S \setminus \{s^*\}$. As soon as $s^*$ gives a better consequence for an act than for another one, the first act is preferred. The other possible states of the world are negligible compared to $s^*$ and only considered when the dominant state does not allow the discrimination between the acts. The decision is then based on the next plausible state...

---

[2]maximal means here that no other partition of $S$ refining this one has the same property

## 5 CONCLUSIONS

The results obtained in Section 4 are rather negative since they underline the poor descriptive ability of models satisfying both QI and the quasi-transitivity of preferences. For this reason, we have investigated possible weakening of our initial axiomatic framework.

A first idea is relaxing axiom S3, arguing that the preference order on consequences may actually depend on the state of the world considered. However, using S2 we can soundly[3] define, for each state $s_j \in S$, a preference relation $\succsim_j$ on $X$ by setting:

$$\forall x, y \in X, \ x \succsim_{P_j} y \iff (f_x \succsim f_y)_{\{s_j\}}$$

Hence, from each relation $\succsim_{P_j}$ we can define a preference relation $\succsim_j$ on $X^S$ by setting:

$$\forall f, g \in X^S, f \succsim_j g \iff f(s_j) \succsim_{P_j} g(s_j)$$

Thus, under QI, linking relation $\succsim$ to the profile $(\succsim_1, \ldots, \succsim_n)$ is still an ordinal aggregation problem with similar difficulties (see Sen(1986)). Thus, cancelling S3 does not really provide new possibilities.

A second possibility is to relax partially the quasi-transitivity of $\succsim$ on $X^S$. In practical applications, most of the acts in $X^S$ are purely imaginary and perhaps not even feasible. Hence, one can be tempted to require quasi-transitivity only on feasible acts. However, using example 3, we could show that this does not provide new possibilities as long as there exists a feasible subset of acts $X'^{S'}$ where $X' \subseteq X$ contains at least 3 distinct elements and $S' \subseteq S$ contains at least 3 non-null states. We could also consider a relaxation of quasi-transitivity for *acyclicity*, i.e., $\forall f_0, \ldots, f_k \in X^S, ((\forall i = 1, \ldots, k, f_{i-1} \succ f_i) \implies \text{not}(f_k \succ f_0))$. Indeed, acyclicity is sufficient to soundly define a choice function selecting the best elements among any non empty subset of $X^S$:

$$\forall Y \subseteq X^S, \ C(Y) = \{f \in Y : \forall g \in Y, \text{not}(g \succ f)\}$$

However, we have also investigated this relaxation and proved that qualitative decision models compatible with acyclicity have also very special features. For example, they necessarily provide some state with an absolute veto, thus limiting the role of other states.

The last possibility is to relax axiom QI which turns out to be a very demanding condition. If we are interested in developing a purely ordinal framework for decision under uncertainty, the most natural weakening of QI is to cancel the "independence" side of the condition while keeping the "ordinal" side. A good solution for this is to assume that the preference $f \succsim g$ not only depends on preferences $f(s) \succsim_P g(s)$, for all $s$, but also on preferences of type $f(s) \succsim_P h(s)$, $g(s) \succsim_P h(s)$, $h(s) \succsim_P f(s)$, $h(s) \succsim_P f(s)$ for some $h$ in $X^S$. This solution has been proposed in the framework of multicriteria Decision Making so as to overcome difficulties due to Arrow-like theorems (see e.g. Perny 1992, Bouyssou and Perny 1992). In the framework of qualitative decision theory, it should be useful to follow the same path and to investigate non-binary preferences. As a first set of examples, consider the following relations:

$$f \Delta_h^+ g \iff [f \succsim h] \succsim_L [g \succsim h]$$
$$f \Delta_h^- g \iff [h \succsim g] \succsim_L [h \succsim f]$$

where $h \in X$ is a reference point within $X^S$ used to specify the aspiration levels of the DM, for each state of the nature. The idea underlying these rules is to compare acts by evaluating their relative abilities in satisfying or missing the DM aspirations. Another interesting set of examples is given by the following qualitative dominance relations:

$$f \Delta^+ g \iff \forall x \in X, [f \succsim f_x] \succsim_L [g \succsim f_x]$$
$$f \Delta^- g \iff \forall x \in X, [f_x \succsim g] \succsim_L [f_x \succsim f]$$

---

[3]S2 can be seen as an independence condition allowing to derive one-dimensional preference relations from $\succsim$. For more details see e.g. Fishburn and Wakker (1995)

These relations must be regarded as natural qualitative counterparts of the so-called *stochastic dominance* relation used in decision under probabilistic uncertainty (see Levy (1992) for a survey on this topic).

Clearly, all these relations are transitive but not necessarily complete. When $\succsim$ is constructed from one of these rules, unanimity holds but *not the QI axiom* and the theorems of this paper do not apply. These examples open new perspectives for designing purely qualitative preference models without QI. This is left for further investigation.

Further work also includes a parallel Savage-like axiomatic study of models based on the comparison of the consequences of the acts in the most plausible states (see (Boutilier 1994; Tan and Pearl 1994; Brafman and Tennenholtz 1996, 1997)). These models do not require preference over consequences and uncertainty to be commensurate and do not obey the QI requirement, but the associated decision rules, when they are explicit, also focus on a decisive set of states: the set of most plausible states. Our approach may also be compatible with generalized qualitative probabilites (Lehman (1996)), that are *incomplete* orders of uncertainty and that require both the transitivity of the indifference relation and unanimity properties stronger than ours. Moreover, generalized qualitative probabilities encompass a notion of neglectibility between events whose links with our notion of decisiveness should be studied.